\pdfoutput=1

\documentclass{article}
\usepackage[hyphens]{url}
\usepackage{microtype}
\usepackage{graphicx}
\usepackage{subfigure}
\usepackage{booktabs} %

\usepackage[breaklinks=true]{hyperref}

\usepackage[accepted]{icml2024}

\usepackage{amsmath}
\usepackage{amssymb}
\usepackage{mathtools}
\usepackage{amsthm}

\theoremstyle{plain}
\newtheorem{theorem}{Theorem}[section]

\theoremstyle{definition}
\newtheorem{definition}[theorem]{Definition}

\theoremstyle{remark}

\usepackage[nameinlink,capitalize,noabbrev]{cleveref}

\crefformat{section}{\S#2#1#3} 
\crefformat{subsection}{\S#2#1#3}
\crefformat{subsubsection}{\S#2#1#3}

\crefname{theorem}{Definition}{Definitions}

\usepackage[textsize=tiny]{todonotes}

\usepackage[inline]{enumitem}
\newenvironment{narrow-itemize}%
{\begin{itemize}%
		\setlength{\itemsep}{0.9pt}%
		\setlength{\parskip}{0.9pt}%
		\setlength{\topsep}{0.9pt}}%
	{\end{itemize}}

\newenvironment{narrow-enumerate}%
{\begin{enumerate}[label=(\arabic*)]%
		\setlength{\itemsep}{0.9pt}%
		\setlength{\parskip}{0.9pt}%
		\setlength{\topsep}{0.9pt}}%
	{\end{enumerate}}

\usepackage[T1]{fontenc}

\begin{document}
	
	\twocolumn[
	\icmltitle{Position: Key Claims in LLM Research Have a Long Tail of Footnotes}

\icmlsetsymbol{equal}{*}

\begin{icmlauthorlist}
\icmlauthor{Anna Rogers}{equal,it}
\icmlauthor{Alexandra Sasha Luccioni}{equal,hf}

\end{icmlauthorlist}

\icmlaffiliation{it}{IT University of Copenhagen}
\icmlaffiliation{hf}{Hugging Face, Canada}

\icmlcorrespondingauthor{Anna Rogers}{arog@itu.dk}
\icmlcorrespondingauthor{Sasha Luccioni}{sasha.luccioni@hf.co}

\icmlkeywords{Machine Learning, Large Language Models, Natural Language Processing}

\vskip 0.3in
]

\printAffiliationsAndNotice{\icmlEqualContribution}

\begin{abstract}
Much of the recent discourse within the ML community 
has been centered around Large Language Models (LLMs), their functionality and potential --- yet not only do we not have a working definition of LLMs, but much of this discourse relies on claims and assumptions that are worth re-examining. We contribute a definition of LLMs, critically examine five common claims regarding their  properties (including `emergent properties'), and conclude with suggestions for future research directions and their framing.
\end{abstract}

\section{Introduction}

Large Language Models (LLMs) have become ubiquitous in the Machine Learning (ML) research landscape, %
and they already impact the lives of thousands of people in contexts ranging from health~\cite{graber2023world,harrer2023attention} to education~\cite{kasneci2023chatgpt}. %
Yet despite the many research articles on LLMs, their very definition remains unclear, and much of this work is based on claims that are often stated, but remain debatable in terms of their framing, theoretical grounding, or empirical evidence. When we, as researchers, repeat such claims uncritically, we contribute to the narratives shaped by business interests \cite{McKelveyDandurandEtAl_2023_News_coverage_of_artificial_intelligence_reflects_business_and_government_hype_not_critical_voices}, and may mislead the public and ourselves. %

\textbf{This position paper argues that LLM research should be more precise with its key terms and claims}. To that end, 
we propose a definition for the term ``LLM'' (\cref{sec:definition}), and we critically examine five common claims about LLM functionality, drawing heavily on both empirical studies and socio-technical critiques of LLMs (\cref{sec:factcheck}). We then consider the impact that these claims have on ML research (\cref{sec:fieldchange}). We conclude with concrete proposals for maintaining rigor and diversity in ML research and practice (\cref{sec:waysforward}).

\section{What Counts as an Large Language Model?} \label{sec:definition}

The common technical definition of ``language models'' is ``models that assign probabilities to upcoming words, or sequences of words in general'' \citep[][p.32]{JurafskyMartin_2020_Speech_and_Language_Processing_3rd_ed_draft}. Yet the term ``large language models'' is currently used in quite a different way, both within the research community (e.g. in the call for papers of academic conferences such as EMNLP~\citeyearpar{emnlp2023}), in news articles~\cite{roose2023does}, and even by legislators at U.S. Senate hearings~\cite{wapocongress2023}). Despite its ubiquity, this definition is far from clear. For instance, how many parameters should a neural network have to qualify as an LLM? And do only Transformer-type architectures qualify? What about multimodal models, such as text-to-image or image-to-text? 

Given the many subfields and contexts in which this term is already used, it is not feasible to impose a single definition and expect everybody to adhere to it. %
But even if there is no agreement, both academic and general public discussions would be more productive if the authors of research papers spelled out or cited their own working definitions. What follows is our own attempt, which we hope could be useful to others (either for using our version, or as a base to be modified by other researchers to reflect their own understanding of this term). It relies upon three definitional criteria: %

\begin{narrow-enumerate}
\item \textbf{LLMs model text\footnote{We understand `text' as `a unit of language in use' \cite{HallidayHasan_2013_Cohesion_in_English}, the product of using that language. %
We see this as a more accurate description, because LLMs model the corpora they are trained on, rather than language in general \cite{Veres_2022_Large_Language_Models_are_Not_Models_of_Natural_Language_They_are_Corpus_Models}.} and can be used to generate\footnote{Most current LLMs produce text in some way, and so our definition focuses on these. To extend it to energy-based models or discriminative models like ELECTRA \cite{ClarkLuongEtAl_2019_ELECTRA_Pre-training_Text_Encoders_as_Discriminators_Rather_Than_Generatorsa}, we could say `can be used to generate or score text'.} it} based on input context, i.e. by selecting the tokens that are the most likely given the partial context provided as input (either masked\footnote{While BERT-style \cite{devlinbert2018} encoders could be used to generate text, it is admittedly a tortuous way of doing so, compared to autoregressive models. However, in both cases we fundamentally solve a classification problem over the model's vocabulary, and autoregressive models could be viewed as a special case of masked language models.} or as a prompt). The text can be in any modality --- characters, pixels, audio, etc.

\item \textbf{LLMs receive large-scale pretraining}, where `large-scale' refers to the pre-training data rather than the number of parameters. The exact threshold for LLM-qualifying volume of data is necessarily arbitrary, and for an English corpus we propose setting it to 1B tokens\footnote{The 1B threshold could need adjustment for different languages and tokenization schemes, if empirical evidence justifies it. We propose to consider raw source data, without any augmentation or considering multiple training runs. We are assuming that these data points would be mostly unique, since it is common practice to deduplicate training data. 

When it comes to multimodal LLMs, the amount of \textit{textual} information in multimodal data can still be approximated via token count (e.g. in transcripts). The other information would be extralinguistic, and a unit for that is yet to be developed -- but compute or parameter counts do not do it justice either. Ideally we would be able to semantically chunk the multimodal content at least as crudely as tokens chunk text, and relate it to the text that it grounds. E.g. we should be able to distinguish between a voiceover over blank screen, a dialogue captured with a fixed camera, or the same dialogue captured from various angles, based on who is speaking.}~(inspired by \citet{billionwords}).

\item \textbf{LLMs are used for transfer learning}, on the assumption that they encode information that can be leveraged in other tasks. Currently the most common transfer learning methods with LLMs are fine-tuning, as in BERT~\cite{devlinbert2018}, and prompting, as in GPT-3~\cite{BrownMannEtAl_2020_Language_Models_are_Few-Shot_Learners}, but there are many other methods  \cite{PanYang_2010_Survey_on_Transfer_Learning,RamponiPlank_2020_Neural_Unsupervised_Domain_Adaptation_in_NLP-A_Survey,AlyafeaiAlShaibaniEtAl_2020_Survey_on_Transfer_Learning_in_Natural_Language_Processing,ZhuangQiEtAl_2021_Comprehensive_Survey_on_Transfer_Learning}.
\end{narrow-enumerate}

According to the above criteria, BERT~\cite{devlinbert2018} and its derivatives do qualify as LLMs, as do models from the GPT series~\cite{radford2018improving}. So do n-gram language models, given that they are derived from a sufficiently large corpus, such as Google Books \cite{lin2012syntactic}. Earlier word-level representations such as word2vec \cite{MikolovChenEtAl_2013_Efficient_estimation_of_word_representations_in_vector_space} are ruled out on the first criterion, when they are viewed by themselves, but their training is conceptually very similar to a masked language model, and they can also use large volumes of text (over 100B tokens for the original word2vec). Modern LLMs can also be used standalone, or for creating representations used in other systems (e.g. BERT's [CLS] token representation \cite{devlinbert2018} fed into classifiers). On our criteria, such representations would not be LLMs, but they are derived from LLMs.

Our first criterion does not rule out multimodal models like GPT-4~\cite{openai2023gpt4}), as long as they output text -- even if they also accept or output images or other modalities. The ``text'' is typically human-written text in a natural language, but it could also be synthetic data (e.g. text created with templates from knowledge base data). The training data of modern LLMs typically also includes text that is not natural language data: code, ascii art, midi music, math notation, chess transcripts etc. While such data can be learned via token prediction, it is not the focus of our definition. %

Our second criterion allows for the inclusion of models such as tinyBERT \cite{jiao2020tinybert}: it has only 4.4M parameters, but its training dataset (via model distillation) is the same as that of the full BERT model, which contains 3.3B tokens.  Our choice of linking the ``large'' part of LLMs to the volume of data rather model size also helps to deal with another edge case: the models that were reduced in size (e.g. via distillation, such as \citet{sanh2019distilbert}) for the sake of computational efficiency, but maintain comparable performance to the original models.

Our third criterion  applies to the ``general-purpose'' LLMs that are purported to be domain-independent. But the core criterion is transfer learning, which may also take place within a specific domain. For example, SciBERT~\cite{beltagy2019scibert} or Galactica~\cite{TaylorKardasEtAl_2022_Galactica_Large_Language_Model_for_Science} are still LLMs, even though they were primarily trained on scientific literature, because they are expected to be used flexibly for a range of tasks within that domain. 

At present, in most cases the three criteria listed above correspond to Transformer-based models that are used to generate text. But having a more concrete definition helps to ground the scientific discourse, and provide a point of reference for updating it in the future. %

As we learn more about both LLMs and neural architectures, our proposed threshold for ``large'' may change, e.g. if there is empirical evidence or theoretical guarantees that a certain volume of natural texts provides sufficient signal for a defined set of linguistic ``skills'' for a given model architecture. 
Hopefuly in the future we would also have more proofs, theoretical rationales, or empirical evidence, based on which the ``large'' part could be further qualified by numerous factors relevant to the performance of the final model: the diversity in the textual data (domains, languages, registers etc.), benchmark contamination, acceptable levels of data augmentation or explicit instruction in the form of annotated data, ratio and role of non-linguistic data, duplicate and near-duplicate data points, allowed number of model runs over the training data, etc.

\vspace{-6pt}

\paragraph{LLMs vs ``foundation'' and ``frontier models''.} LLMs are also sometimes referred to as ``foundation models'', a term proposed by Bommasani et al~\citeyearpar{bommasani2021opportunities} to refer to ``any model that is trained on broad data (generally using self-supervision at scale) that can be adapted (e.g., fine-tuned) to a wide range of downstream tasks''. This partly corresponds to our criteria (2) and (3), but makes no attempt to quantify their scale. It is also intentionally broader, so as to include e.g. models for computer vision or protein data. The term ``LLM'' is more specific, and useful in studies modeling language data.

Another recently proposed term is ``frontier models'', defined as ``highly capable foundation models that could exhibit sufficiently dangerous capabilities'' \cite{AnderljungBarnhartEtAl_2023_Frontier_AI_Regulation_Managing_Emerging_Risks_to_Public_Safety}. This relies on the above ``foundation model'' term, and only adds the criterion of ``sufficiently dangerous capabilities'', which the authors acknowledge to be vague. Various kinds of risks from LLMs are beyond the scope of this work, but see \cref{ref:emergence} for a relevant discussion of ``emergent properties''.

\section{Fact-checking LLM Functionality} \label{sec:factcheck}

We discuss five common claims about LLMs: that LLMs are robust (\cref{sec:robust}), that they systematically achieve state-of-the-art results (\cref{subsec:llm-sota}), that their performance is predominantly due to their scale (\cref{subsec:scale}), that they are ``general-purpose technologies (\cref{subsec:gpt}) and that they exhibit emergent properties (\cref{ref:emergence}). We are not saying that all these claims are completely false -- but they all have many caveats, which are mentioned much less frequently. %
By collecting existing evidence and counter-arguments, we aim to highlight some of the gaps and inconsistencies in our current knowledge and to help 
orient future work so as to address these gaps.

\subsection{Claim: LLMs are Robust}
\label{sec:robust}

Early symbolic AI approaches are often described as ``brittle'' because of their strict dependence on pre-formulated knowledge and lack of robustness outside of the distribution they were trained on. \citet[p.1175]{lenat1988thresholds} described this as ``\textit{a plateau of competence, but the edges of that plateau are steep descents into complete incompetence}''. With the advent of LLMs, the issue of robustness is seen to be much less prominent. For instance, \citet[p.109]{bommasani2021opportunities} state: ``\textit{pretraining on unlabeled data is an effective, general-purpose way to improve accuracy on [out-of-distribution] test distributions}''. LLMs are often presented as multi-task learners that are robust without explicit supervision, even outside the distribution they were trained on~\cite{radford2019language,hendrycks2019using,hendrycks2020pretrained}.

Indeed, we have overcome the problem of the \textit{steep} descents into complete incompetence: unfamiliar inputs no longer completely break the system. But deep-learning-based ML systems are still fundamentally brittle, only in a different way: according to \citet[p.3]{Chollet_2019_On_Measure_of_Intelligence}, they are \textit{``unable to make sense of situations that deviate slightly
from their training data or the assumptions of their creators''}. Chollet goes even further, stating in an interview that there are no fixes for this issue~\cite{heaven2019deep}. 
Impressive as the latest LLMs are, %
they still make errors %
even in simple tasks like adding numeric literals ~\cite{chang2023language}. Moreover, they are just as vulnerable to adversarial attacks \cite{zou2023universal} as earlier models \cite{WallaceFengEtAl_2019_Universal_Adversarial_Triggers_for_Attacking_and_Analyzing_NLP}.

One could argue that ``robust'' does not mean ``perfect'' -- but in that case, what does it mean? For an ML engineer, it is something like ``sufficiently useful in practice''. From that point of view, two situations are possible: the model will be deployed in the conditions either (a) guaranteed to be similar to its training distribution in all aspects that matter for its performance, or (b) expected to diverge from that. \textbf{Our current LLM-based solutions may be sufficiently robust for in-distribution scenarios, but few would argue the same for out-of-distribution cases.} Case (b) covers many, if not most, LLM application areas: text classifiers will continually encounter new topics and domains, language usage will evolve, the correct answers to factual questions will change, discourse strategies for interaction with AI-enabled chatbots will shift, people will adapt what they post online and to the privacy and surveillance concerns, etc. And failures of ML systems may have real-world consequences for those who diverge the most from its core distribution: e.g. people may be denied asylum due to errors of machine translation systems, something that we have already seen happen~\cite{nalbandian2022eye}.  %

One well-studied cause of brittleness in the LLMs of the BERT generation was shortcut learning \citep[][inter alia]{mccoy2019right,rogers2020getting, branco2021shortcutted,choudhury2022machine} -- models picking up undesirable spurious correlations from the training data, which are very likely to exist in all the larger datasets used by the data-hungry deep learning systems \cite{gardner2021competency}. This problem is still there for the latest LLMs: when they fail on counterfactual tasks or adversarial perturbations, this suggests that their successes are due not to learning the general principles behind a certain operation, but some narrow heuristic that does not transfer to new contexts. \cite{wu2023reasoning}.

In the few-shot evaluation paradigm, we also now have a new robustness problem. The art of `prompt engineering' \cite{liu2023pre} arose out of the prompt sensitivity phenomenon: slight variations in the phrasing of the prompt that would not make much difference to a human can lead to very different LLM output \cite{lu2021fantastically,pmlr-v139-zhao21c}. In a recent evaluation of 30 LLMs, \citet[p.12]{liang2022holistic} conclude that ``all models show significant sensitivity to the formatting of prompt, the particular choice of in-context examples, and the number of in-context examples across all scenarios
and for all metrics''. The reports of sensitivity to exact wording keep coming for the latest models, including GPT-4~\cite{LeeBubeckEtAl_2023_Benefits_Limits_and_Risks_of_GPT4_as_AI_Chatbot_for_Medicine,GanMori_2023_Sensitivity_and_Robustness_of_Large_Language_Models_to_Prompt_Template_in_Japanese_Text_Classification_Tasks}.%

\subsection{Claim: (Few-shot) LLMs Are State-of-the-Art} \label{subsec:llm-sota}

LLM-based approaches have become the default in the current research literature, and are largely perceived to be the current SOTA (state-of-the-art) across NLP benchmarks. For example, \citet[p.179]{gillioz2020overview} state: ``\textit{Models like GPT and BERT relying on this Transformer architecture have fully outperformed the previous state-of-the-art networks. It surpassed the earlier approaches by such a wide margin that all the recent cutting edge models seem to rely on these Transformer-based architectures.}''

The above was written in the days of fine-tuned Transformer-based LLMs like BERT. At this point, such a statement needs to be considered in the context of the distinction between \textit{few-shot performance} (ostensibly out-of-domain performance achieved by a model that was not specifically trained on a given task), vs performance of a \textit{model fine-tuned for a given task}. For example, both BERT \cite{devlinbert2018} and GPT-3 \cite{BrownMannEtAl_2020_Language_Models_are_Few-Shot_Learners} were presented with evaluation on question answering, among other tasks, but the former was fine-tuned, while the latter evaluated in a few-shot way. 

Generally speaking, an ML model that has been trained on some domain data can be reasonably expected to perform better in that domain than a comparable model that hasn't received such training. This means that we now have two different notions of SOTA, where the few-shot setting could reasonably be expected to yield worse performance vs the same model if it was fine-tuned, but requires less data and training. Still, the current research papers introducing LLMs often include only few- or zero-shot evaluations, which creates the impression that this is the only evaluation that matters. For example, OPT~\cite{zhang2022opt} was 
evaluated on 16 tasks concurrently without fine-tuning, establishing new accuracy in several of them; the same goes for models such as PaLM~\cite{chowdhery2022palm}, LLaMa~\cite{touvron2023llama1}, and many others. 

We broadly agree that pre-trained models based on Transformer architecture are likely to be the current SOTA when they are provided additional in-domain supervision. But \textbf{most of the current LLM evaluation discourse shifted to few- or zero-shot evaluation, and in that context the SOTA claim may not hold. Hence, many of the current results may not actually represent the current SOTA}.

When we consider direct comparisons between few-shot LLMs and supervised systems, not based on the same LLMs, the winner 
depends on the specific case, but the few-shot LLM is not at all guaranteed to win -- especially in the ``true few-shot'' setting, where prompts are not selected based on extra held-out data \cite{PerezKielaEtAl_2021_True_FewShot_Learning_with_Language_Models}. They may also be at disadvantage in the niche domains or tasks like sequence labeling that are less straightforward to formulate as a text generation task. Consider that few-shot GPT-3 \cite{BrownMannEtAl_2020_Language_Models_are_Few-Shot_Learners} is on the SuperGLUE \cite{WangPruksachatkunEtAl_2019_SuperGLUE_Stickier_Benchmark_for_General-Purpose_Language_Understanding_Systems} leaderboard with the average score of 71.8, compared to a score of 84.6 achieved by a fine-tuned RoBERTa model~\cite{LiuOttEtAl_2019_RoBERTa_Robustly_Optimized_BERT_Pretraining_Approach}. As another example, recent work on NER \cite{wang2023gptner} and relation extraction \cite{wan-etal-2023-gpt} explicitly formulated their problem as few-shot learning generally trailing behind supervised approaches in their tasks of interest, and their contribution - as overcoming that (in both cases, with the help of a supervised method in the pipeline).   %
\citet{openai2023gpt4} claimed that GPT-4 outperforms unspecified fine-tuned models on 6 out of 7 verbal reasoning tasks, but provided no detail on model or benchmark selection, making it impossible to reproduce or verify these results.

One more consideration for the ``(few-shot) LLMs are SOTA'' statement is that it implies a direct competition with other methods, with which a meaningful comparison is possible. But what are we comparing -- the model architectures or training data? ML as a scientific field focuses on the former, but most LLM leaderboards present apple-to-orange comparisons.

Finally, for both few-shot and fine-tuned evaluation of LLMs, \textbf{most of the reported results should be taken with a grain of salt because of test data contamination}. For example, GPT-4 received a lot of press coverage due to the claim of achieving a score that falls in the top 10\% of test takers on a simulated bar exam \cite{KatzBommaritoEtAl_2023_GPT4_Passes_Bar_Exam} -- but that result was soon questioned  on grounds of improper evaluation and possible data contamination \cite{Martinez_2023_ReEvaluating_GPT4_Bar_Exam_Performance}. 

In the GPT-3 report, OpenAI itself documented how hard it is to avoid benchmark contamination~\cite{BrownMannEtAl_2020_Language_Models_are_Few-Shot_Learners}. By now, multiple studies presented evidence of the presence of common NLP benchmarks in multiple datasets used for training LLMs~\cite{dodge2021documenting,magar2022data,blevins2022language}, which can inflate LLM performance in certain tasks and datasets. The LM Contamination Index\footnote{\url{https://hitz-zentroa.github.io/lm-contamination/}}, a collaborative effort to document benchmark contamination, currently has 375 entries for various benchmarks and models across different tasks. Furthermore, a recent study has documented the effect where GPTs score higher on the ``old'' benchmarks than on the new ones \cite{liu2023evaluating}, which strongly suggests that the previously reported evaluation results may be inflated.

\subsection{Claim: (LLM) Scale Is All You Need} \label{subsec:scale}

Scaling has played a central role in the success of LLMs -- starting with the `scaling laws' paper for causal language models~\cite{kaplan2020scaling}, which found that their performance improves with scaling the  model size, data size, and also the amount of compute used for training. This analysis was subsequently expanded to other benchmarks, modalities and downstream tasks~\cite{ghorbani2021scaling,alabdulmohsin2022revisiting,hernandez2021scaling,hoffmann2022training}. It is often mentioned as a key factor\footnote{To be fair, the focus on scaling does not entail that other factors are completely irrelevant, and we are not saying that the entire ML community believes that ``scale is all you need''. But it is also fair to say that scaling has received a lot more attention than other factors, in particular data, which has long been considered as a less prestigious kind of work \cite{SambasivanKapaniaEtAl_2021_Everyone_wants_to_do_model_work_not_data_work_Data_Cascades_in_High-Stakes_AI}.} in LLM performance; for instance, \citet[p.1]{huang2022large} state: ``\textit{scaling has enabled Large Language Models (LLMs) to achieve state-of-the-art performance on a range of Natural Language Processing (NLP) tasks}''. The focus on scaling is in line with the ``bitter lesson'' of \citet{sutton2019bitter}, which states that we should stop working on methods based on the human knowledge of the target problem, and embrace ``search and learning'', because ``the only thing that matters in the long run is the leveraging of computation''. 

Indeed, the scaling hypothesis seems to be supported by the fact that LLMs have been growing in size for several years: e.g. BERT-base in 2018 had 340M parameters~\cite{devlinbert2018}, and in 2022 PaLM had 540B parameters in 2022~\cite{chowdhery2022palm}. And there is evidence that, even with the same architecture and training data, larger models tend to perform better, even with adversarial evaluation \cite{BhargavaDrozdEtAl_2021_Generalization_in_NLI_Ways_Not_To_Go_Beyond_Simple_Heuristics,RayChoudhuryRogersEtAl_2022_Machine_Reading_Fast_and_Slow_When_Do_Models_Understand_Language,wang-etal-2022-super}.

However, it is important to keep in mind that many of the best-known LLMs scaled both the number of parameters and their training data concurrently (besides any differences in architecture and training set-up)\footnote{E.g. BERT was trained on roughly 3.3 billion tokens, and PaLM's training data had 780 billion, representing a growth of 1500 times in terms of model size and 260 times in terms of dataset size. This came with a big improvement in performance: BERT achieved an accuracy of 70.1\% on the \href{https://aclweb.org/aclwiki/Recognizing_Textual_Entailment}{RTE dataset}, PaLM achieves an accuracy of 95.7\%, with possible data contamination.}. %
While this seems to support the scaling laws hypothesis as formulated by \citet{kaplan2020scaling}---we do not know which of these components is most responsible for the improvement, and most LLMs are not directly comparable by more than one of these criteria.

Furthermore, simply increasing the size of the training dataset has a complex relation with its quality. %
When we train LLMs on cleaner, more diverse text data, we do not merely provide more \textit{data}, but more \textit{knowledge} (by better covering the kinds of information that may be required for performing the model's task). Then, \textbf{the improved performance likely results from the fact that we are supplying more knowledge, and not just more scale/computation, and it is bounded by availability of such knowledge.} %
Unlike in chess and Go, for LLMs the new knowledge has to come from manually created sources, which do not cover all scenarios that may arise in practice, and hence computation can only take us as far as the data goes.\footnote{This is not to say that such a system cannot be useful in some cases, or that it cannot exhibit some generalization within a space sufficiently covered by data: the linguistic fluency of the current LLMs is a testament to the possibility of learning ``closed'' systems such as syntax.} 
In practice, we break the central tenet of the ``bitter lesson'' all the time: when we identify an area where a model under-performs, a common solution is to try to collect, label, synthesize, or even create more data for that problem, which is tantamount to injecting manually-curated knowledge. %

This would be in line with the fact that the developers of high-performing LLMs now spend more effort on improving the quality of the training data. %
E.g. both PaLM reports \cite{chowdhery2022palm,anil2023palm} devote considerable effort to data cleaning, with PaLM 2 explicitly attributing higher performance on English benchmarks to higher quality data %
\citep[p.9]{anil2023palm}; the paper accompanying the Chinchilla model also makes a similar ``quality over quantity'' point with regards to data~\cite{hoffmann2022training}. The Llama 3 announcement\footnote{\url{https://ai.meta.com/blog/meta-llama-3/}} stresses a heavy investment in pre-training data. The key result of Phi model \cite{gunasekar2023textbooks} is also explicitly presented as a smaller high-performing model, made possible by training on higher-quality data.

Furthermore, the last few years have seen an increased skepticism around the scaling hypothesis, starting with `efficient scaling' proposals for Transformer models, which showed that smaller, more efficient models can outperform bigger ones in certain settings~\cite{tay2021scale}. This was further explored in practice via the Inverse Scaling Prize, showing that there are tasks, such as logical reasoning and pattern matching, where the performance does not seem to improve with model size~\cite{mckenzie2022inverse}. %

Finally, let us consider the fact that there are high-performing  open-access models such as LLaMa series~\cite{touvron2023llama1,touvron2023llama2}, which perform very well despite being much smaller than GPT-3. The success of techniques such as knowledge distillation~\cite{pan2020meta,sanh2019distilbert} and sparsity~\cite{srinivasan2023training} also strongly suggests at least that the model size by itself is not the `secret sauce'. And, of course, it can be a deal-breaker for  deploying models in production, irrespective of  performance gains. %

\subsection{Claim: LLMs Are General-Purpose Technologies} \label{subsec:gpt}
According to \citet[p.3]{eloundou2023gpts}, \textit{Generative Pre-trained Transformers (GPTs) are general-purpose
technologies (GPTs)}. This framing can be found both in the media~\cite{forbes2023, forbes2023_2} and preprints ~\cite{tamkin2021understanding,liu2023generate}.\footnote{The preprints by \citet{eloundou2023gpts,liu2023generate} were co-authored by researchers affiliated with OpenAI.}

GPT (General-Purpose Technology) is a term used by economists and historians to refer to technologies that are both era-defining and pervasive over time; however, what qualifies as a GPT and what does not has been hard to delineate, since technologies are often nested within other systems of recursive technologies and systems~\cite{knell2022tools}. 
A widely-accepted definition by economists Lipsey and Carlaw proposes 4 criteria for a technology to be considered general-purpose: (1) it is a single, recognisable generic technology, (2) it comes to be widely used across the economy, (3) it has many different uses and (4) it creates many spillover effects~\cite{lipsey2005economic}. According to these criteria, a total of 24 technologies such as the wheel, the printing press and electricity are considered GPTs.

At this point, it is hard to assert the general-purpose status of LLMs by the above criteria. They are not a single, generic technology (as discussed in~\cref{sec:definition}). They are currently not widely used in different domains~\cite{bekar2018general, prytkova2021ict}, and remain auxiliary tools even in those domains where they are most used~\cite{bianchini2020deep}. In terms of it usage, LLMs are not widely used in the vast majority of economic activities, and they are reliant upon specific commodities such as large amounts of GPUs and highly specialized labor, which are only available in a small number of industries and by a handful of organizations~\cite{bresnahan2019artificial}. A further constraint is the fact that LLMs, like all deep learning-based technology, can be expected to work best in-distribution -- and it is unclear to what extent issues with robustness (\cref{sec:robust}) will limit its broader utility in the vast number of economic sectors associated with the smaller communities and languages. 

As for the final criterion, regarding the potential spillover effects of these technologies, it is really too early to tell what these may be~\cite{bresnahan2019artificial,crafts2021artificial,natale2020imagining}. The overnight popularity of ChatGPT, which is often seen as proof of the widespread usage of LLMs in broader society, can mainly be attributed to the creation of a simple user interface on top of models that had been previously available via APIs~\cite{eloundou2023gpts}, rather than technological novelty.
\textbf{Until we can meaningfully assess the adoption and application of LLMs, we should refrain from putting them in the same conceptual category as the printing press}, and focus on defining what they can and cannot be used for (and under what conditions). %

\subsection{Claim: LLMs Exhibit ``Emergent Properties''}
\label{ref:emergence}

LLMs are often discussed in terms of their ``emergent properties''.\footnote{Some researchers discuss ``emergent abilities'' rather than ``properties''. Both of these terms could also use better definitions. Our interpretation is that they are used interchangeably in LLM research, and the chief difference is the anthropomorphizing framing for ``ability''. The underlying construct in case of \citet{WeiTayEtAl_2022_Emergent_Abilities_of_Large_Language_Models} seems to be NLP ``tasks'', on the assumptions that these tasks have construct validity, and specific evaluation datasets provide valid measurements of performance on these tasks. The task/benchmark confusion is exacerbated by the fact that in BIG-Bench \cite{srivastava2023imitation} the constituent datasets are sometimes named as if they were tasks (e.g. ``IPA transliterate''). 
} %
Among such properties, various researchers have included few-shot learning \cite{bommasani2021opportunities}, ``augmented prompting'' techniques such as ``chain of thought'' \cite{WeiWangEtAl_2022_ChainofThought_Prompting_Elicits_Reasoning_in_Large_Language_Models}, predicting intermediate computation results \cite{NyeAndreassenEtAl_2021_Show_Your_Work_Scratchpads_for_Intermediate_Computation_with_Language_Models} or whether the answer is correct \cite{KadavathConerlyEtAl_2022_Language_Models_Mostly_Know_What_They_Know}, and, on the input side -- instruction following \cite{OuyangWuEtAl_2022_Training_language_models_to_follow_instructions_with_human_feedback,WeiBosmaEtAl_2021_Finetuned_Language_Models_are_ZeroShot_Learners}. %

In line with the confusion about the term ``LLM'' (\cref{sec:definition}), the term ``emergent properties'' seems to be used in at least 4 distinct ways in current LLM research:

\begin{definition}
\label{def:emergence-unseen}
A property that a model exhibits despite not being explicitly trained for it. E.g. \citet[p.5]{bommasani2021opportunities} refers to few-shot performance of GPT-3 \cite{BrownMannEtAl_2020_Language_Models_are_Few-Shot_Learners} as ``an emergent property that was neither specifically trained for nor anticipated to arise''.
\end{definition}

\begin{definition}
\label{def:emergence-seen}
(Opposite to \autoref{def:emergence-unseen}): a property that the model learned from the pre-training data. E.g. \citet[p.8]{deshpande-etal-2023-honey} discuss emergence as evidence of ``the advantages of pre-training''.
\end{definition}

\begin{definition}
\label{def:emergence-size}
A property ``is emergent if it is not present in smaller models but is present in larger models.''  \citep[p.2]{WeiTayEtAl_2022_Emergent_Abilities_of_Large_Language_Models}.  
\end{definition}

\begin{definition}
\label{def:emergence-sharp} A version of \cref{def:emergence-size}, where what
makes emergent properties ``intriguing'' is ``their sharpness, transitioning seemingly instantaneously from not present to present, and their unpredictability, appearing at seemingly unforeseeable model scales''~\citep[p.1]{schaeffer2023emergent}
\end{definition}

\autoref{def:emergence-seen} seems describe the expected outcome of successful training. In this sense, ``emergent properties'' could be referred to simply as ``learned properties''.

\autoref{def:emergence-size} is questionable if the ``emergent'' behavior can be achieved in smaller models\footnote{See e.g. \citet{SchickSchutze_2021_It_Not_Just_Size_That_Matters_Small_Language_Models_Are_Also_FewShot_Learners,gao-etal-2021-making}. One could object that these studies provide the smaller models a lot of extra help (reformatting the inputs, selecting extra models etc.) This is true, but the commonly reported few-shot performance is also not really few-shot, and reliably choosing good prompts becomes \textit{harder} with larger models (presumably due to more complex decision boundaries)  \cite{PerezKielaEtAl_2021_True_FewShot_Learning_with_Language_Models}.}. %
Even more importantly, we still have the contamination problem: if both the bigger and smaller models were trained on data similar to the test data, the bigger one would still be reasonably expected to perform better just because it has more capacity to learn it.
In that case, \autoref{def:emergence-size} follows from the \autoref{def:emergence-seen}.

With respect to \autoref{def:emergence-sharp}, \citet{schaeffer2023emergent} %
make a convincing case that such sharp increases in performance may be an artifact of the chosen evaluation metric\footnote{\citet{Wei_2023_Common_arguments_regarding_emergent_abilities} and \citet[p.38]{AnderljungBarnhartEtAl_2023_Frontier_AI_Regulation_Managing_Emerging_Risks_to_Public_Safety} argue that this is not important, because ``non-smooth'' metrics used for real tasks are the ones we care about. But then the question remains whether this is something special about LLMs, or just a property of the metric. Simple models also have such ``emergent properties'', as shown by \citet{schaeffer2023emergent} for a shallow nonlinear autoencoder.%
} 
rather than a fundamental property of scaling the model.

An even bigger issue with \autoref{def:emergence-sharp} is that we simply do not have enough data points to say that the increase in performance is sharp: e.g. if we had intermediate model sizes between the commonly-used 13B, 70B, and 150+B, we would likely see a smooth transition. \citet{Wei_2023_Common_arguments_regarding_emergent_abilities} acknowledges that, but argues that
the ``emergence'' phenomenon is still interesting if there are large differences in predictability: for some problems, performance of large models can easily be extrapolated from performance of models 1000x less in size, whereas for others, even it cannot be extrapolated even from 2x less size. But the cited predictability at 1,000x less compute refers to the GPT-4 report \cite{openai2023gpt4}, where \textit{the developers knew the target evaluation in advance}, and specifically optimized for ``predictable scaling''. This is in contrast with the unpredictability at 2x less compute for \textit{unplanned} BIG-Bench evaluation by \citet{WeiTayEtAl_2022_Emergent_Abilities_of_Large_Language_Models}. %

So, we are left with \autoref{def:emergence-unseen}, which can be interpreted in two ways: 
\begin{definition} \label{def:emergence-hard}
A property is emergent if the model was not exposed to training data for that property.
\end{definition}

\begin{definition}\label{def:emergence-soft}
A property is emergent even if the model was explicitly trained for it -- as long as the developers were unaware of it.
\end{definition}

Per \autoref{def:emergence-soft}, it would appear that we are training LLMs as a very expensive method to discover what data exists on the Web. For example, the fact that ChatGPT can generate chess moves that are plausible-looking (but often illegal)\footnote{\scriptsize \url{https://reddit.com/r/AnarchyChess/comments/10ydnbb/i_placed_stockfish_white_against_chatgpt_black/}} is to be expected, given the vast amount of publicly-available chess transcripts on the Web.

Per \autoref{def:emergence-hard}, %
we can prove that some property is emergent only by showing that there was no evidence that could have been the basis for the model outputs in the training data. For commercial models with undisclosed data such as ChatGPT, this is out of the question. But we would go further and argue that the emergent properties according to \autoref{def:emergence-hard} are only a hypothesis (if not wishful thinking) even for the ``open'' LLMs, because %
so far we are lacking detailed studies (or even a methodology) to consider the exact relation between the amount and kinds of evidence in the training text data for a particular model output. Hence, to the best of our knowledge, \textbf{there is no evidence for the existence of ``emergent properties'' per \autoref{def:emergence-hard}}. Until we have such evidence, it seems strange for the ML community to conclude that the best explanation for high performance is not-learned-from-data emergent properties---especially in the face of evidence to the contrary.\footnote{E.g. \citet[p.12]{liang2022holistic}, evaluate 30 LLMs and conclude that ``regurgitation (of copyrighted materials) risk clearly correlates with model accuracy''. \citet{liu2023evaluating} report that ChatGPT and GPT-4 perform better on older compared to newly released benchmarks, and \citet{mccoy2023embers} show that their performance depends on probabilities of output word sequences in web texts. \citet{lu2023emergent} show that ``emergent abilities'' of 18 LLMs can be ascribed mostly to in-context learning. For in-context learning itself, the results of \citet{ChanSantoroEtAl_2022_Data_Distributional_Properties_Drive_Emergent_In-Context_Learning_in_Transformers} suggest that it happens only in Transformers trained on sequences, structurally similar to the sequences in which in-context learning would be tested.}

What about benchmarks that are newly created and tested on for the first time in a given study -- aren't they, by definition, uncontaminated? We argue not: in the absence of a methodology to even compare test and training data beyond trivial exact matches, ``new'' tests may be so similar to something observed that they do not really count as ``new''. This has been a known problem even with benchmark datasets, the size of which is very small compared to LLM training data\footnote{E.g. \citet{lewis-etal-2021-question} found that about 30\% of test samples in 3 popular QA benchmarks have near-duplicates in train data.}. The LLM training data may also include something that is not even public\footnote{In particular, the OpenAI models could have been trained not only on Web data, but also on the data of thousands of researchers who over the past year submitted their trickiest test cases to GPT-3 API (the policy for the API data to be opted out of training by default only changed in Spring 2023).} on the internet, and thus not easily searchable to confirm the contamination.  

Perhaps the most striking example of how unreliable the ``new'' test examples are is the ``sparks of intelligence'' study \cite{bubeck2023sparks}. Using the methodology of newly constructed test cases, checked against public web data, and their perturbations, \citet{bubeck2023sparks} notably concluded that GPT-4 possesses ``a very advanced theory of mind''. At least two studies have since come to the opposite conclusion \cite{sap2023neural,ShapiraLevyEtAl_2023_Clever_Hans_or_Neural_Theory_of_Mind_Stress_Testing_Social_Reasoning_in_Large_Language_Models}. 

The above discussion focused on the way the term ``emergence'' is currently used in LLM research. In philosophy of science, there are many nuanced discussions of emergence to which we cannot do justice here, but broadly it can be characterized as a phenomenon in complex systems that is dependent on its constituent parts, but is also  distinct from them. The leading example in Stanford Encyclopedia of Philosophy is a tornado: it consists of dust and debris, but it ``its features and behaviors appear to differ in kind from those of its most basic constituents''\cite{OConnor_2020_Emergent_Properties}. Emergent phenomena are described by a different scientific field, on a different level than the constituent phenomena: it is possible to understand the behavior of tornadoes without understanding particle physics. Does this notion of emergence apply in the context of ML?

Our take is that such a notion of emergence would hold for LLMs: the information they encode is not explainable by the model weights, at least at the current state of interpretability research. However, this also applies to most deep learning models (e.g. a simple MLP for sentiment classification), and the latest LLMs are not something special. %

\section{How Does This Change the Theory and Practice of ML?} \label{sec:fieldchange}

The success of LLMs brought our field an increase in funding, real-world applications, and attention. But the perception of their robustness (\cref{sec:robust}) and SOTA status (\cref{subsec:llm-sota}), the idea that their success is purely due to scale (\cref{subsec:scale}), that they are a general-purpose technology (\cref{subsec:gpt}), and their ill-defined  ``emergent properties'' (\cref{ref:emergence}) also contribute to the following trends:

\begin{narrow-itemize}
\item \textit{Homogeneity of approaches.} If LLMs are so great, why would we pursue anything else? It is understandable that most current NLP research is focused on LLMs, but this also means that as researchers we have most of our eggs in one basket, and become less likely to develop alternatives or see the gaps in our knowledge.
\item \textit{De-democratization.} \citet{openaiWIRED} estimates that training GPT-4 cost more than \$100 million; although the true number is unknown, the increase in computational requirements of LLMs is clear~\cite{thompson2022computational}. This means that graduate students, independent researchers, and even most academic labs struggle to either reproduce existing results or train new LLMs that would require large amounts of compute.
\item \textit{Industry influence.} Recent research into the affiliations of researchers in ML~\cite{abdalla2021grey,ahmed2020democratization,birhane2022values,whittaker2021steep} has shown both a steep increase in industry presence in conference publications over the past years (e.g. a 180\% growth for NLP~\cite{abdalla2023elephant}), and an influence over the topics of research being pursued by the community, such as robustness and similar challenges, instead of more theory-oriented topics. Given the above point, this trend makes sense, since only industry labs with extensive funding can afford to train and deploy LLMs. But it has implications for the diversity of ideas and approaches in the field.
\item \textit{(Further) decreased reproducibility.} Reproducibility was an issue in ML even before LLMs \cite{Crane_2018_Questionable_Answers_in_Question_Answering_Research_Reproducibility_and_Variability_of_Published_Results,cohen2018three,pmlr-v97-bouthillier19a}, and it has not gotten much better~\cite{belz2021systematic,belz2022quantified,belz2022metrological}. LLMs pose additional issues both in reproducing their training and fine-tuning \cite{SellamYadlowskyEtAl_2021_MultiBERTs_BERT_Reproductions_for_Robustness_Analysis,McCoyMinEtAl_2019_BERTs_of_feather_do_not_generalize_together_Large_variability_in_generalization_across_models_with_similar_test_set_performance} and inference results \cite{hagmann2023inferential}, not to mention the lack of control over API-only models that can change over time \cite{chen2023chatgpts,Rogers_2023_Closed_AI_Models_Make_Bad_Baselines}. %
If the tested model is deprecated%
, or changes in any way (e.g. by extra training or fine-tuning, changes in its associated parameters, filtering mechanisms or system prompts), %
the results reported in the study will no longer be reproducible. They might not even hold at the time of publication -- and we will have no idea why. %

\end{narrow-itemize}

The ideas that LLMs are robust (\cref{sec:robust}) and exhibit emergent properties (\cref{ref:emergence}) further contribute to the impression of irrelevance of any theory of the linguistic, social, cognitive, or any other phenomena that LLMs are supposed to model. This is unsatisfactory if the goal is scientific research, but even from a purely engineering perspective this stance is dangerous, since it entails that we either cannot or do not need to provide specifications for cases where a given model is safe or unsafe to use. As a result, LLMs may be deployed in society at large, without compelling evidence of their performance in all the target scenarios, or among the demographic groups that are likely underrepresented in the training data \cite{Bender2021OnTD}, or across the less-resourced languages \cite{zhu2023multilingual,bang2023multitask,lai2023chatgpt,huang2023languages,ziems2023multivalue}. %

\section{Ways Forward} \label{sec:waysforward}

We have argued that there are many deep knowledge gaps in LLM research, and the field is changing in ways that make these gaps less likely to be addressed. We now conclude with some concrete recommendations for future work. 

\textit{Maintaining diversity of research approaches.} We advocate not for stopping all work on LLMs, but for maintaining a healthy diversity of approaches and tasks. Among other things, this means efforts by the conferences to ensure fair reviews for the ``niche'' submissions.\footnote{E.g. papers on ``niche'' topics can have priority in reviewer assignments \cite{rogers-etal-2023-report}.} Putting all of our eggs in the proverbial LLM basket runs the risk of missing out on new research directions and exciting opportunities to make significant connections with other fields, such as linguistics and cognitive science. %

\textit{Defining terminology.} We discussed the lack of clarity on the very term ``large language model'' (\cref{sec:definition}), as well as ``emergent properties'' (\cref{ref:emergence}). 
Ideally, at least in research papers we would start with specifying what we mean by a common-but-vague term -- e.g. ``vision-and-language Transformer model'' or ``a Transformer-based autoregressive language model with 200 billion parameters''. 
Although it is impossible to control what terminology is used in popular science or journalism about LLMs, as experts and researchers we have a responsibility to be clear and precise when discussing our domain of expertise (see ~\citet{lacroix2023ethics}). 

\textit{Not using ``closed'' models as baselines.} Since the launch of GPT-4 in March 2023, numerous studies have used it both as a benchmark to compare different methods and  models~\citep[e.g.][]{liu2023evaluating,sun2023chatgpt} as well as an object of scientific study in itself~\citep[e.g.][]{bubeck2023sparks,WeiTayEtAl_2022_Emergent_Abilities_of_Large_Language_Models, zhang2023exploring}. We believe that this is problematic for several reasons:

\begin{narrow-itemize}
\item It may produce inflated performance reports due to undisclosed training data, which could be contaminated with benchmark data. This could then create the false impression that the ``closed'' model is intrinsically so much better that there is no competing with it.
\item It normalizes methodologically dubious apples-to-oranges comparisons to black box models. 
\item The reproducibility of evaluations is significantly reduced (see \cref{sec:fieldchange}).  
\item Where the ``closed'' models are provided commercially, academic researchers essentially perform free work to help the company improve their product -- and they even pay\footnote{According to one US academic researcher we spoke to, the monthly spending on OpenAI API in their lab is capped at \$10K a month, and otherwise would sometimes even exceed that sum.} for that privilege, often with funding from public grants that could be spent on improving the public resources. Moreover, if the general perception is that only this kind of work constitutes ``frontier'' LLM research, the labs without the financial means to pay for the API access are at severe disadvantage, and the field gets homogenized even further. 
\end{narrow-itemize}
While we recognize that some analyses of proprietary, ``closed'' models like GPT-4 and PaLM can be useful (e.g. audits, red-teaming), relying on these models as baselines, and especially expecting others to do so, is both unfair and unreliable. We recommend using open-source or at least open-access\footnote{We do not advocate for using \textit{only} open-source models, because the models available now can be ``open'' in different relevant ways (code, weights, license, training data), and these aspects matter more/less for different research questions and real-world applications~\cite{solaiman2023gradient}. A model like Mistral \cite{jiang2023mistral} is more transparent than GPT-4 in terms of its architecture, but not the training data. A model like BLOOM \cite{le2023bloom} is more transparent in terms of training data and can be an excellent object of research, but it is not strictly open-source, e.g. because of its RAIL license. Our position is that in research, we should aim to use the most transparent options available, and thankfully there seems to be a trend towards more openness in this sense (e.g. the recent OLMO model \cite{groeneveld2024olmo} and its training dataset DOLMa \cite{soldaini2024dolma}).} models like FLAN-T5~\cite{flan2022}, BLOOM \cite{scao2022bloom}, LLaMa~\cite{touvron2023llama1,touvron2023llama2} and FALCON~\cite{falcon40b}. 

\textit{Further rigorous studies of LLM functionality.} There are numerous knowledge gaps about LLM fuctionality. What exactly makes a given type of model (e.g. a Transformer-based LLM) SOTA on a given task, given that there are no confounds such as differences in training data and model capacity? What kinds of brittleness do LLMs exhibit, and how to mitigate that? Do they really have any emergent properties (and how are these defined)? How can we ensure specific kinds of robustness? 
Large-scale benchmarks \citep[e.g.][]{srivastava2023imitation} do not address these questions, since they are constructed on the basis of data availability rather than definitions of specific phenomena \cite{RajiDentonEtAl_2021_AI_and_Everything_in_Whole_Wide_World_Benchmark}. 
These questions are not answered by the current large-scale evaluation endeavors \citep[e.g.][]{liang2022holistic}. While they are very valuable, there are too many confounding variables in the published models, and more extensive error analysis and exploration is needed to disentangle them. 
Access to both models and their training data is especially important for this purpose, since it can help establish links between LLM behavior and their training data \cite{piktus2023roots} and understand their biases~\cite{gururangan2022whose,johnson2022ghost, abid2021persistent}.

\textit{Developing better evaluation methodology.} There has been progress towards multi-dimensional evaluation that takes into account more than performance \cite{EthayarajhJurafsky_2020_Utility_is_in_Eye_of_User_Critique_of_NLP_Leaderboards,liang2022holistic,ml-energy-leaderboard} and reproducibility \cite{DodgeGururanganEtAl_2019_Show_Your_Work_Improved_Reporting_of_Experimental_Results,ulmer-etal-2022-experimental,magnusson-etal-2023-reproducibility}, but much more remains to be done. There is also a dire need for structural incentives to encourage reproducibility efforts and publication of negative results. And most importantly, we need a lot more work on the validity of the underlying constructs \cite{RajiDentonEtAl_2021_AI_and_Everything_in_Whole_Wide_World_Benchmark,Schlangen_2021_Targeting_Benchmark_On_Methodology_in_Current_Natural_Language_Processing_Research}. Ideally, LLM evaluation would target specific types of language processing rather than broadly defined tasks  \cite{RibeiroWuEtAl_2020_Beyond_Accuracy_Behavioral_Testing_of_NLP_Models_with_CheckList,RogersGardnerEtAl2023}. %
It would carefully explore the decision boundaries and specific kinds of brittleness~\cite{KaushikHovyEtAl_2019_Learning_Difference_That_Makes_Difference_With_Counterfactually-Augmented_Data,GardnerArtziEtAl_2020_Evaluating_NLP_Models_via_Contrast_Sets}, and control for potential confounds so as to focus purely on model architecture. As discussed in \cref{ref:emergence}, we also need new methodology for systematically linking the model outputs to potential evidence in the training data at LLM scale -- relevant current efforts include the work on memorization \citep[][\textit{inter alia}]{thakkar2021understanding, carlini2022quantifying, chang2023speak} and providing search indices for LLM training data \cite{piktus2023roots,MaroneVanDurme_2023_Data_Portraits_Recording_Foundation_Model_Training_Data}. All this is in parallel to the general problems with evaluating open-ended generation, including the dependence on generation strategies~\cite{meister2021language}, and issues with popular metrics such as perplexity~\cite{wang2022perplexity}.

\section{Conclusion}
As researchers, we cannot help but observe the near-universal focus on LLMs in recent years and their impact on our field.  This paper discusses several knowledge gaps and common claims that should be taken with a grain of salt, as well as the ways in which LLMs changed the research landscape. We argue for more rigor in definitions, experimental studies, and evaluation methodology, as well as higher standards for transparency and reproducibility. We hope to open the door for more discussion of the contribution of LLMs to ML research, and how we can better leverage their strengths while understanding their limitations.

\section*{Acknowledgements}

We would like to thank all anonymous reviewers of this paper. Their insightful comments were invaluable for sharpening the discussion. Rob van der Goot, Christian Hardmeier, Yacine Jernite, Margaret Mitchell, Dennis Ulmer read the early versions of this paper and provided feedback. We also thank Ryan Cotterell, Ishita Dasgupta, Laura Gwilliams, Julia Haas, Anna Ivanova, Tal Linzen, Ben Lipkin, Asad Sayeed for their insights and discussion.

This work was partly supported by DFF Inge Lehmann grant to Anna Rogers (3160-00022B).

\section*{Impact Statement}
This work is a position paper, expressing the personal views of its authors, and supported by evidence and arguments we cite rather than new experimental work. Our goal is not to criticize a specific research direction or technology as a whole, but to raise the awareness about the issues with the lack of clarity with key terms and claims in our field.

We recognize that we come from a position of privilege, holding positions at institutions located in the Global North. Our opinions do not reflect the lived experiences of visible minorities or those who come from less privileged institutions and geographical regions. We have endeavored to have our manuscript read by colleagues from other domains of expertise and other institutions, but we recognize that it does not represent the experiences and perspectives of all members of the ML community.

\bibliography{custom}
\bibliographystyle{icml2024}

\end{document}